# Adaptive Distraction Context Aware Tracking Based on Correlation Filter

Fei Feng[1], Xiao-Jun Wu[1, *], Tianyang Xu[1, 2], Josef Kittler[2], Xue-Feng Zhu[1]
[1]School of Internet of Things Engineering, Jiangnan University, Wuxi, China.
[2]Centre for Vision, Speech and Signal Processing (CVSSP), University of Surrey, Guildford, UK
xiaojun_wu_jnu@163.com

## Abstract

*The Discriminative Correlation Filter (CF) uses a circulant convolution operation to provide several training samples for the design of a classifier that can distinguish the target from the background. The filter design may be interfered by objects close to the target during the tracking process, resulting in tracking failure. This paper proposes an adaptive distraction context aware tracking algorithm to solve this problem. In the response map obtained for the previous frame by the CF algorithm, we adaptively find the image blocks that are similar to the target and use them as negative samples. This diminishes the influence of similar image blocks on the classifier in the tracking process and its accuracy is improved. The tracking results on video sequences show that the algorithm can cope with rapid changes such as occlusion and rotation, and can adaptively use the distractive objects around the target as negative samples to improve the accuracy of target tracking.*

## 1. Introduction

Visual object tracking is an important research topic in the field of computer vision. In the process of tracking, the target has the problem of scale variations, illumination variations, occlusion, deformation, etc. [1], which makes how to achieve robust tracking becomes a challenge for visual object tracking. After a period of development, the object tracking algorithm also has a large number of public benchmarks, such as OTB2013[1], OTB100[2], Tcolor-128 [3], VOT2016 [4] and so on. The CF algorithm has also achieved high tracking accuracy [5, 24, 25] on the open database and is also in the leading position in various object tracking competitions [4, 21, 22].

The CF algorithm is a typical discriminative algorithm. Multiple classifiers are used to distinguish between the target and the background, so as to achieve the goal of object tracking [23, 26]. The CF algorithm uses circulant convolution to construct a large number of training samples for classifier training, and uses correlation operations to determine the target position. The appearance of the target in the current frame is used as a training sample to update the classifier. The complexity of the computation is reduced by the Fourier transform, in order to achieve high computation efficiency. However, the CF algorithm still has some deficiencies. It uses a single model in the training of the classifier [20]. When the target is subject to occlusion

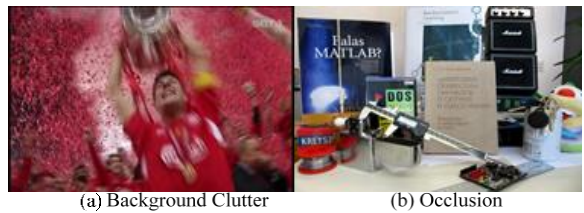

(a) Background Clutter   (b) Occlusion

Figure1 Impact on Target Accuracy in Object Tracking

and rotation as shown in Figure 1(b), the classifier may fail, and because the training samples are generated by a circulant matrix, the target near the edge of the search area can easily be considered as a negative sample, resulting in a tracking failure. In the detection of the target, the image block similar to the target is easily misidentified as the target position due to the correlation property as shown in Figure 1(a), which affects the tracking accuracy.

There are many object tracking algorithms that do not use the CF framework. For example, Lu et al. combined the sparse representation with the adaptive structured local sparse appearance model (ASLA) [13]. ROSS et al. proposed incremental visual tracking (IVT) algorithm [14]. Zhang et al. proposed the Compressive Tracking (CT) [15].

In this paper, we propose an adaptive distraction context aware tracking method is used to find similar distraction block around the target. It is used as a negative sample in the process of training the classifier, which can avoid tracking failures caused by high similarity between the image block and the target. Thus, the proposed adaptive distraction context aware correlation filter tracking algorithm considers the information of the response map in the target detection process, and selects negative sample image blocks to train the classifier with the aim to improve classification accuracy. The method avoids considering an occluding object or background clutter as the target, and improves the tracking accuracy.

## 2. Correlation Filter

CF tracking algorithms have been used for many years. In the early days, Bolme et al. proposed a minimum output sum of squared error filter (MOSSE) tracking algorithm [6], and applied the concept of correlation filter in tracking for the first time. Henriques et al. introduced circulant matrix on the basis of MOSSE and proposed Kernel Correlation Filter (KCF) [7] to improve computational efficiency while using Histogram of Oriented Gradient (HOG) instead of the original gray features. Danelljan et al. and Lukezic et al. proposed the spatially regularized correlation filters (SRDCF) [8] and the discriminative correlation filter with channel and spatial reliability (CSRDCF) [9], It solves the problem of scale change of the target and of unnatural boundary samples generated by the circulant matrix. Danelljan et al. and Bertinetto et al. introduced color features in the CF framework, and proposed color names (CN) features [10] and complementary learners for real-time tracking algorithms (Staple) [11]. Matthias et al. proposed a context-aware correlation filter tracking (DCFCA) [12] algorithm to increase filter discrimination by including more background information during the training.

The CF algorithm performs a cyclic shift operation on the base sample $x = (x_1, x_2, ..., x_n)$, which generates a large number of virtual training samples and uses these virtual training samples (negative samples) and base samples (positive samples) to learn the classifier $f(x) = w^T x$. In order to minimize the error function between the sample $x_i$ and its corresponding ridge regression target $y_i$, the objective function is expressed as[7]

$$\min_w \sum_{i=0}^{n-1} \|w^T x_i - y_i\|_2^2 + \lambda \|w\|_2^2 \quad (1)$$

where, $w$ is the filter, and $\lambda$ is a regularization parameter. The optimization function is the standard ridge regression problem. The closed form solution is given as

$$w = (X^T X + \lambda I)^{-1} X^T y \quad (2)$$

where, each row vector in the sample data matrix $X$ is obtained by a circular shift of the first row of the base sample $x$. Each element in $y$ is the ridge regression target value $y_i$ for the corresponding sample $x_i$, and $I$ is an identity matrix. The circulant matrix $X$ has the following properties [10]:

$$\begin{aligned} X &= F diag(\hat{x}) F^H \\ X^H &= F diag(\hat{x}^*) F^H \end{aligned} \quad (3)$$

where $x$ is the first row vector of the circulant matrix, $(\bullet)^*$ is the complex conjugate of vector $(\bullet)$, $(\hat{\bullet})$ is the Fourier transform of $(\bullet)$, $F$ is the discrete Fourier transform matrix, $H$ is the Hermitian matrix, which is defined as $F^H = (F^*)^T$. Equation (3) indicates that the convolution operation of the original data sample $x$ can be computed in the frequency domain. Diagonalizing the sample $x$, and multiplying it with the Fourier matrix $F$ and $F^H$ is to quickly obtain the sample matrix after the cyclic shift.

Using the property of Equation (3), the circulant matrix $X$ is diagonalized, and Equation(2) can quickly learn the model filter by the Fourier transform [16].

$$\hat{w} = \frac{\hat{x} \odot \hat{y}}{\hat{x}^* \odot \hat{x} + \lambda} \quad (4)$$

When the tracking algorithm detects the target position in a new frame, a convolution operation is performed on the trained filter $w$ and the candidate image $z$ obtains a response map $R$, where $R$ represents the circulant matrix generated by the image $z$, and the position of the maximum value in $R$ is the position of the target in a new frame.

$$R(w, z) = Zw \Leftrightarrow \hat{R} = \hat{z} \odot \hat{w} \quad (5)$$

where Z is the circular matrix of the candidate image.

## 3. Distraction Context Aware Tracking Based on Discriminative Correlation Filter

During the tracking of the CF algorithm, the image content around the target ha an impact on the tracking results. By using the background information surrounding the target to train the classifier, the discrimination ability of the classifier can be improved, thereby improving its tracking accuracy. Therefore, we propose an adaptive distraction context aware tracking algorithm that uses the response map obtained during the detection of the algorithm to adaptively find similar image content around the target. The presence of these similar images can lead to tracking inaccuracy. Adding them as negative samples into the training set can sharpen the separability of the target appearance from the confusing background, and help to suppress false detections, so that the real position of the target can be detected accurately during the detection process.

In frame $t$, the circulant matrix $Z$ of the candidate image is convolved with the model $w_{t-1}$ from the frame 1 to frame $t-1$ to obtain the target position in response $R$. There commonly are multiple peaks in the response map, and these peak locations are called interest points. In the traditional CF algorithm, the location of the maximum peak in the response map is the real position of the target. Based on this principle, it can be assumed that the position corresponding to secondary peaks of the response map cannot be the real position of the target. These images may be distractors or an object that may occlude the target. The images corresponding to these secondary peaks can be identified as strong negative samples and these negative samples come from the classifier model from 1 to $t-1$ frames and the response map of the current image. They represent the contextual information, rather than just random samples, and consequently they are more discriminative. Including them in the training set provides not only more information about the target and background, but also contains labelled negative information that could

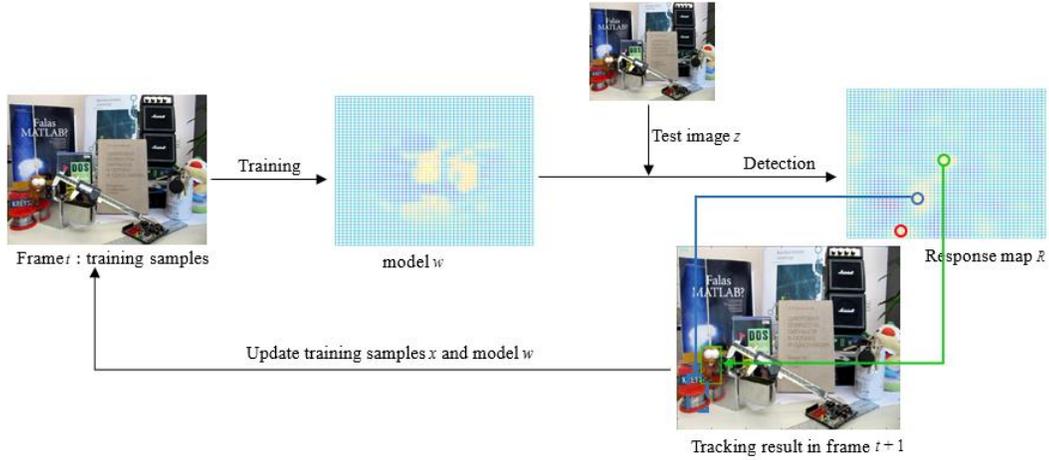

Figure2 Training, detection and adaptive search for interest points process of the proposed algorithm.

distract the classifier. In the training phase, the weights of different training samples are increased to achieve more accurate classification, and adaptively solve the challenges of tracking, caused by occlusion and background clutter. For a clearer understanding of the algorithm, Figure 2 shows the stages of training, detection and adaptive search for interest points carried out by the proposed algorithm. At time $t$, we first extract features from the training sample $x$. From the response map $R$ in Figure2, we can see that there are multiple interest points. The position with the maximum response in the map is considered as the target position, as shown by the green circle. There are other interest points in the response map $R$, as shown with the blue and red circles. The blue circles, close to the target, may affect the tracking results. Adding data corresponding to these interest points to the training set as negative samples is expected to enhance the discriminatory power of the model.

More specifically, we add the images of the $k$ interest points to the training set to build classifier, $w$. Recalling the traditional CF algorithm optimization formula (Equation (1)), as the distraction image blocks $d_i, i \in k$ are added as negative samples, we can write down a new optimization formula

$$\min_w \sum_{i=0}^{n-1} \|w^T x_i - y_i\|_2^2 + \lambda_1 \|w\|_2^2 + \lambda_2 \sum_{i=1}^{k} \|w^T d_i\|_2^2 \quad (6)$$

We want the negative samples to produce as low response to the filter as possible. Since both $x_i$ and $d_i, i \in k$ are image blocks, only the corresponding ridge regression values are different. Therefore, the image blocks in Equation (6) can be constructed as a new image matrix, recorded as $M, M \in \mathbb{R}^{(k+1)n \times n}$. Using a regression label matrix $\bar{y}, \bar{y} \in \mathbb{R}^{(k+1)n}$ to assign ridge regression value labels to different image blocks in this data matrix [12], we can rewrite the Equation (6) as follows.

$$\min_w \sum_{i=0}^{n-1} \|Mw - \bar{y}\|_2^2 + \lambda_1 \|w\|_2^2$$

$$M = \begin{bmatrix} X \\ \sqrt{\lambda_2} D_1 \\ \vdots \\ \sqrt{\lambda_2} D_i \end{bmatrix}, \bar{y} = \begin{bmatrix} y \\ 0 \\ \vdots \\ 0 \end{bmatrix} \quad (7)$$

Where, the data matrix $D$ is circulant matrix obtained by a cyclic shift of its first row vector $d_i$. The circulant matrix $X$ is also obtained by a cyclic shift of its first row vector $x_i$, where each row vector is a sample. Each element in $y$ corresponds to the ridge regression target value $y_i$ of sample $x_i$. Because $D$ contains negative samples for training the classifier, the corresponding ridge regression value tag is 0. Equation (7) is known to be a least-squares problem. Its closed form solution can be obtained by

$$w = (M^T M + \lambda I)^{-1} M^T \bar{y} \quad (8)$$

Since this closed form solution is similar to the solution of the CF algorithm classifier, using the Fourier property of the circulant matrix in Equation (3), Equation (8) is optimized in the frequency domain, and its fast learning classifier formula can be obtained as follows.

$$\hat{w} = \frac{\hat{x} \odot \hat{y}}{\hat{x}^* + \hat{x} + \lambda_1 + \lambda_2 \sum_{i=1}^{k} \hat{d}_i^* \odot \hat{d}_i} \quad (9)$$

The updating strategy of the standard CF algorithm is used in the target model updating step of our algorithm. The specific target model updating formula is $w_t = (1-\theta) \cdot w_{t-1} + \theta \cdot w_t$, where, $\theta$ is the learning rate. It means the target model from 1 to $t$ frames consists of two parts, one is the old model kept from 1 to $t-1$ frames, with a weight of $1-\theta$ and the other one is a new model learned from the $t$-th frame. The target detection step is similar to the detection strategy of the CF algorithm [7], but the online adaptive updating rule [17] is added to make the algorithm more robust. In frame $t$, we use the model from 1 to $t$ frames to replace the model from 1 to $t-1$ frames to compute $R$ in Equation (5).

In the process of extracting features from a sample, if $x$ is a color image, the features of the target are described by HOG (31-dimension) [7] and CN (10-dimension) [10]; If $x$ is a gray image, the feature becomes the Intensity Channel (IC) [18] and HOG [7]. In order to reduce the dimensionality of the data in the training process and reduce the computational complexity, the principal component analysis (PCA) is used to reduce the dimensionality of the features after their extraction from the images. After the dimensionality reduction, the HOG features are 10-dimensional, and CN features are 3-dimensional. Then, using Equation (9) to train the model, the candidate images and the model are convolved by Equation (5) to obtain the target response map.

During the process of tracking and detection, the position of the interest points obtained from the response map, $R$ may not be close to the target, as shown by the red circle in Figure 2. The interest points far from the target have a little influence on the target localisation. They can therefore be ignored without affecting the solution. In practice, we identify them by measuring their distance from the target. Those at a distance exceeding a threshold are excluded from the training set. We only include the images of interest points close to the target, and treat them as negative samples. This interest point selection strategy also improves the computational efficiency of the algorithm. Furthermore, if the response value of candidate interest points is not high enough, they are also not considered as distraction blocks to the target. Only when an interest point response value is higher than 20% of the real target response value, the image block is considered as a relevant interest point. Recording the value of $i$ peaks in the response map $R$ as $f_i$, and ranking them in the descending order, we propose to keep the first three, denoted as $F(i)$, as candidates for inclusion. Among the three responses, $F(1)$ would be considered as the target position, and $F(2)$ and $F(3)$ as distraction images.

$$F(i) = \arg decrease(f_i) \qquad (10)$$

Determining the response value and position of $F(2)$ and $F(3)$, with respect to the target, the points are added to the training set, otherwise they are discarded. The algorithm flow is described in Algorithm 1.

**Algorithm1 Adaptive distraction context aware tracking based on correlation filter (ADCACF)**

**Input:** training image $x$, test image $z$, target position $p_{t-1}$ at time $t-1$.

**For frame=1: end frame**

Initialize all parameters, read video sequence images

1. Train the classifier $w$ for the training samples $x$ by Equation (9);

2. Obtain the response map $R$ and the target position $p_t$ in time $t$ by Equation (5);

3. Find interest points in response map $R$, obtain $F(i)$ by Equation (10);

4. Analyse the position of $F(2), F(3)$;

**if** interest points are around the target, $F(2), F(3) > 20\% \cdot F(1)$

Add the points into training the classifier

**Else** give up this interest point

5. Update model $w$ and training samples $x$;

**end**

**Output:** target position $p_t$ at time $t$

### 4. Experiments

The algorithm is implemented in Matlab and C language for mixed programming. The experimental results are obtained under Matlab2017 and VS2015 software versions. Experiments are performed on the tracking benchmark OTB100 [2] and Tcolor-128 [3] and compared with other algorithms under the same experimental conditions. This section compares the proposed algorithm with (shown in the experiment results as ADCACF) with KCF [7], CSRDCF [9], SRDCF [8], DCFCA [12], Staple [11], CSK [19], ASLA [13], IVT [14] and the CT [15] algorithm.

All the baseline algorithms use the original author's open source implementations and default parameters for comparison experiments. The experiments compare the performance in terms of the success rate (SR), its summary as area under curve (AUC), and speed measured in terms of frames per second (FPS). The success rate (SR) is expressed as the proportion of frames with an overlap ratio greater than 0.5 to the total number of frames. Let the target rectangle be represented by a rectangle $A_T$, and denote the ground-truth of the target by a rectangle $A_G$. The overlap

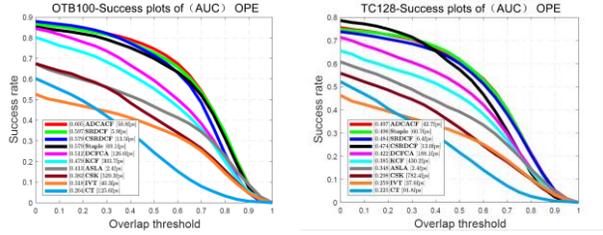

Figure3 the AUC of different algorithms on OTB100 and TC128

ratio is defined as $SR = \frac{area(A_T \cap A_G)}{area(A_T \cup A_G)}$. As SR uses a single data evaluation sequences and does not quantify fairness and representativeness, AUC of success rate is used to provide an alternative measure of performance. FPS is the number of frames per second achieved by each algorithm. Section 4.1 presents the results of an overall analysis of the algorithms on the tracking benchmarks. Section 4.2 presents the performance of the algorithms for video attributes. In Section 4.3 we present a sensitivity analysis of the proposed algorithm to changes to meta parameters.

### 4.1. Experiments on Visual Tracking Benchmark

Figure 3 shows the AUC values achieved by the different algorithms on benchmarks OTB100[2] and Tcolor-128[3] (expressed as TC128 in the chart), representing the success rate of object tracking on these two benchmarks. As the AUC figures of the bottom ranking algorithms are much lower, Table 1 presents the AUC values of only the top 7 algorithms in Fig.3.

Table1 the AUC values of different algorithms on OTB100 and TC128

| Benchmark<br>Algorithms | OTB100 | TC128 | Average FPS |
|---|---|---|---|
| ADCACF | 0.605 | 0.497 | 46.75 |
| Staple | 0.579 | 0.496 | 64.9 |
| SRDCF | 0.597 | 0.484 | 6.15 |
| CSRDCF | 0.579 | 0.474 | 13.25 |
| DCFCA | 0.512 | 0.422 | 157.35 |
| KCF | 0.479 | 0.385 | 366.95 |
| ASLA | 0.413 | 0.348 | 2.4 |

Comparing the AUC results on the OTB100 data set, ADCACF (0.605) ranks the first, with a slightly higher performance than the second algorithm, SRDCF (0.597). However, our algorithm is much faster than SRDCF. These two are then followed by CSRDCF (0.579), Staple (0.579), and DCFCA (0.512). The DCFCA [12] algorithm also suppresses the background around the target. However, their method uses a fixed sampling. In contrast, the suppression mechanism in our method is adaptive and the sampling changes during object tracking. This is reflected in the tracking performance of ADCACF. CSK (0.382), IVT (0.318), and CT (0.264) achieve lower AUC and are ranked the last as they do not consider the possibility of training sample distortion.

In the Tcolor-128 dataset, the ADCACF (0.497) algorithm is also the best performer, slightly better than Staple (0.496). Staple uses the color histogram information for the feature representation to enrich the target's characteristic description. The use of histogram statistics enhances the tracking robustness of the algorithm under the condition of deformation and illumination. In contrast ADCACF suppresses the distractive image blocks, which are similar to the target, in the process of training the filter and thus enhances the discriminative power of the filter. The robustness of the ADCACF algorithm is improved in the case of occlusion, and background clutter.

The last line in Table2 is the average FPS of the algorithms on the two databases. KCF is the fastest at 366.95 frames per second, and the ASLA algorithm is the worst at 2.4 seconds per second. The ADCACF algorithm is ranked 6th among the above 10 algorithms, lower than KCF, CSRDCF, DCFCA, CSK, CT algorithm, but still has 46.75 frames per second.

### 4.2. Attribute-Based Evaluation

Video sequences present different challenging attributes of object tracking, such as occlusion (OCC), scale variations (SV), deformation (DEF), motion blur (MB), and In-plane Rotation/Out-of-plane rotation (IPR/OPR), illumination variations (IV), fast motion (FM), background clutter (BC), out of view (OV), and Low Resolution (LR). Each video sequence contains two or more of these attributes depending on the key changes in the appearance of the target and the scene. The standard video sequences in the OTB100[2] and Tcolor-128[3] are classified into the above categories of video attributes. The object tracking performance of the algorithms for different video attributes is analyzed in Figure 4. It shows the AUC for different video attributes in the OTB100 video set for the 10 algorithms studied. The title shows the video attribute. The parentheses give the number of sequences in the video benchmark that contain this video attribute. The brackets in the figure indicate the speed of tracking in FPS.

It can be seen from the figure that the ADCACF algorithm performs well in background clutter, motion blur, rotation, occlusion, and target out of view. It is the best of the 10 algorithms tested, owing to ADCACF adaptively using similar image content close to the target as negative samples, rather than selecting negative samples at fixed positions. The negative samples are the points of interest in the response map, the position of which is close to the target. Distraction blocks close to the target would have a greater impact on object tracking than the distant blocks. The suppression of such points effectively enhances the discrimination of the filter. The background clutter and

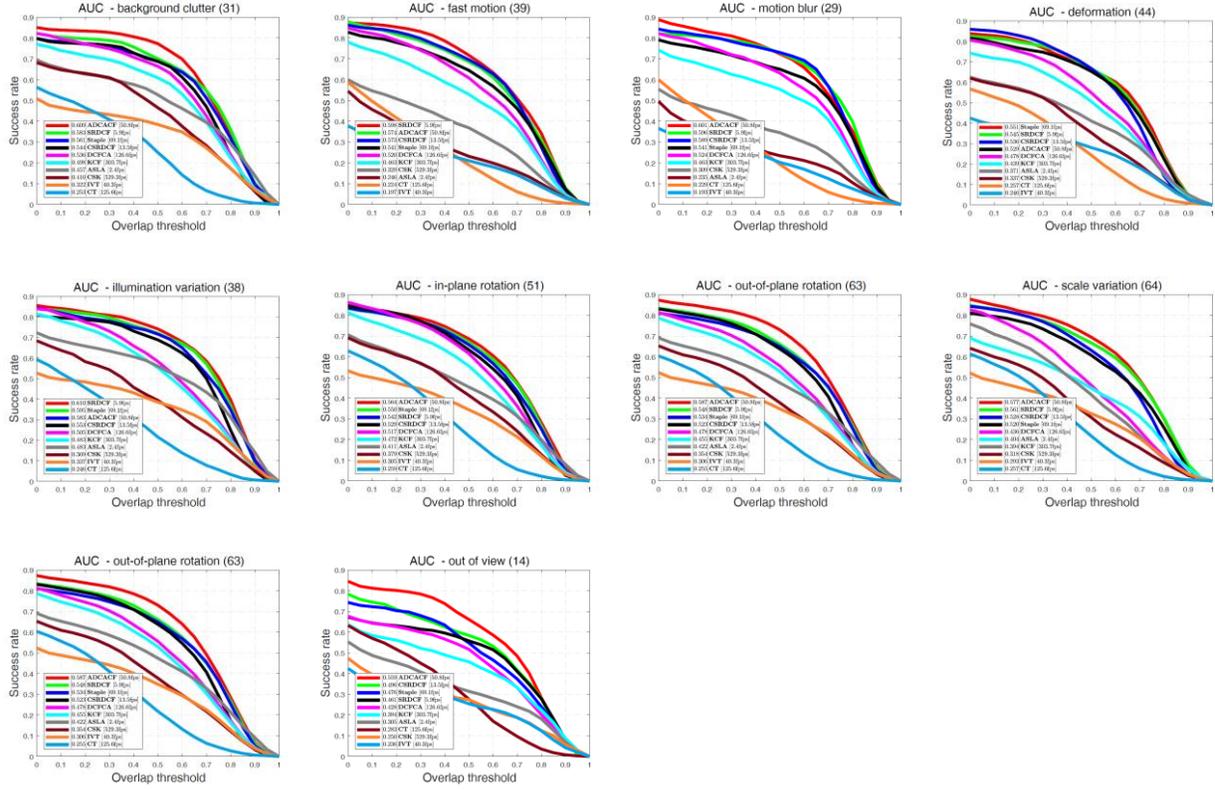

Figure4 Different Algorithms Tracking Results with Different Video Attributes in the OTB100

occlusion may cause a tracking error because of the presence of similar image blocks close to the target. After the adaptive suppression of the negative samples, the classifier exhibits better performance.

It can also be seen from Figure 4 that the ADCACF is less effective in illumination variations, fast motion and deformation, while SRDCF, Staple and other algorithms show good tracking results in these attributes. As ADCACF does not use statistical features such as color histograms, it is more sensitive to illumination variations. Also because the circulant matrix causes the distortion of the training samples, the ADCACF does not add a suppression term to the filter in order to learn the filter as quickly as possible. In consequence, partially distorted training samples interfere with the training of the classifier. This issue will be addressed in the future.

Due to the space limitations, the AUC curve of the algorithms for different video attributes in the Tcolor-128 is given in Appendix I. In those figures, it can also be seen that the ADCACF deals well with the background clutter, rotation, motion blur, and occlusion, but struggles under fast motion and illumination variations. It also shows that the ADCACF adaptively suppresses similar distraction block and has a certain effect on the object tracking.

Figure 5 presents the tracking results of different algorithms on some standard sequences. We selected the top 6 of the algorithms in Figure 3 for comparison. They are ADCACF, SRDCF, CSRDCF, DCFCA, KCF, and Staple respectively. The red line in the figure5 represents the ADCACF, the green one represents the SRDCF, the blue one represents the DCFCA, the black line represents the Staple, the pink one represents the CSRDCF, and the light blue rectangle represents the KCF. It can be seen from the figure that ADCACF performs better that other tracking algorithms in background clutter, rotation, motion blur, and occlusion.

**4.3. Parameter Sensitivity Analysis**

The ADCACF algorithm determines the suppression block, on the basis of the traditional CF algorithm optimization. In addition to the regularization parameter $\lambda_1$, which is the same as in the standard CF method and its value is not adjusted. The parameter $\lambda_2$ is added to the optimization formula (6). These parameters are set as $\lambda_1 = 10^{-2}, \lambda_2 = 15$. The HOG cell size of the ADCACF is 9, and the CN cell size is 4. The results report AUC and distance precision (DP). DP is the percentage of frames whose estimated location is within a given threshold

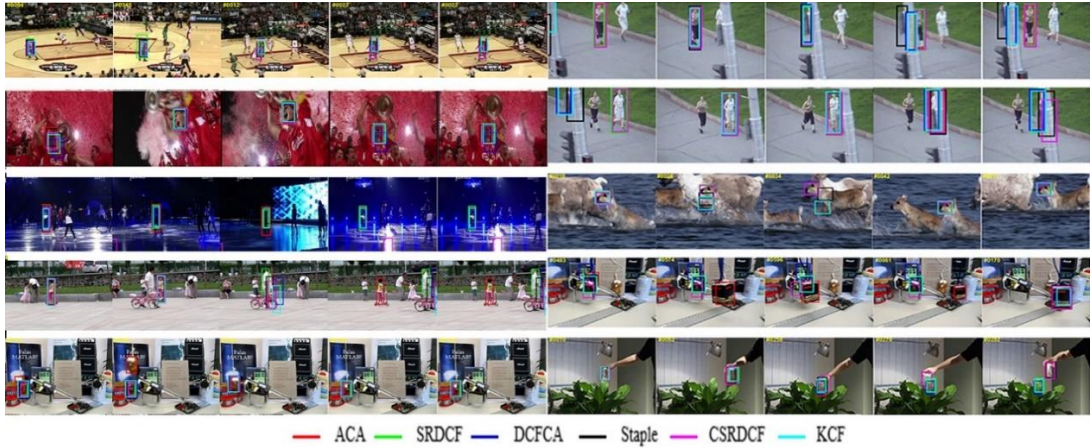

Figure5 Partial video sequence tracking results

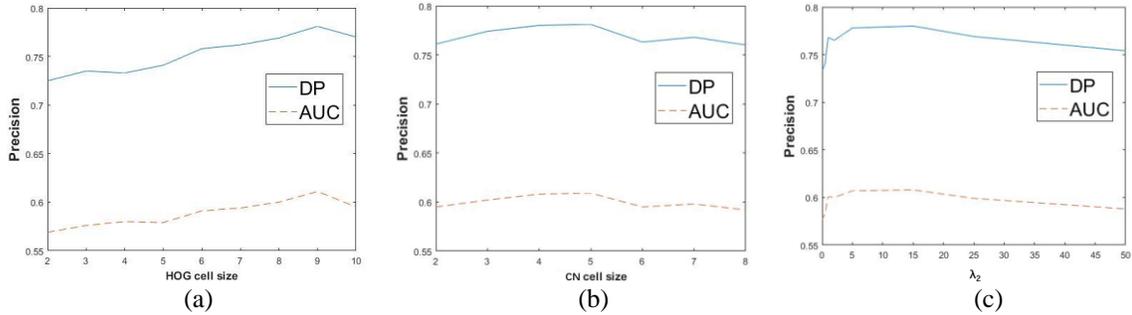

Figure6 Influence of different parameter values on the algorithm distance to the ground truth. In this paper, we choose threshold = 20.

From Figure 6(a), it can be seen that when the cell size of HOG changes from 2 to 10, the DP value of the algorithm fluctuates slightly between 0.72 and 0.78, the AUC fluctuates around 0.58. The change in sample cell size has less effect on the DP and AUC of the algorithm. From Figure 6(b), we can see that when the sampling cell size under the CN changes from 2 to 8, the changes of AUC and DP values are all within 0.05. We can see the change is small and the curve is relatively stable. Therefore, the algorithm is insensitive to changes in the size of the CN sample cell. From Fig. 6(c), we can see that when $\lambda_2$ changes from 0.01 to 50, the DP fluctuates within 0.05, and the AUC within 0.04. Therefore, the algorithm is not very sensitive to changes of $\lambda_2$, and the algorithm has certain degree of robustness.

## 5. Conclusion

In the process of training the classifier, the adaptive distraction-context aware tracking algorithm uses the relevant response value of the previous frame to find the image blocks that are similar to or interfere with the target. Those close to the target are selected as negative samples to suppress the distractive effect of these blocks or of occlusion on the detection process in the next frame. This innovative mechanism increases the robustness of the tracking process and reduces the tracking drift caused by the distractive areas and occlusion. Through extensive experiments on tracking benchmarks, it is shown that the proposed algorithm has good performance in the presence of background clutter, motion blur, rotation, occlusion and the target out of view. Although it improves the tracking accuracy when the target is distracted, the algorithm is not robust enough for tracking in sequences exhibiting target deformation, fast motion and illumination variations. These challenges will be the subject of future research.

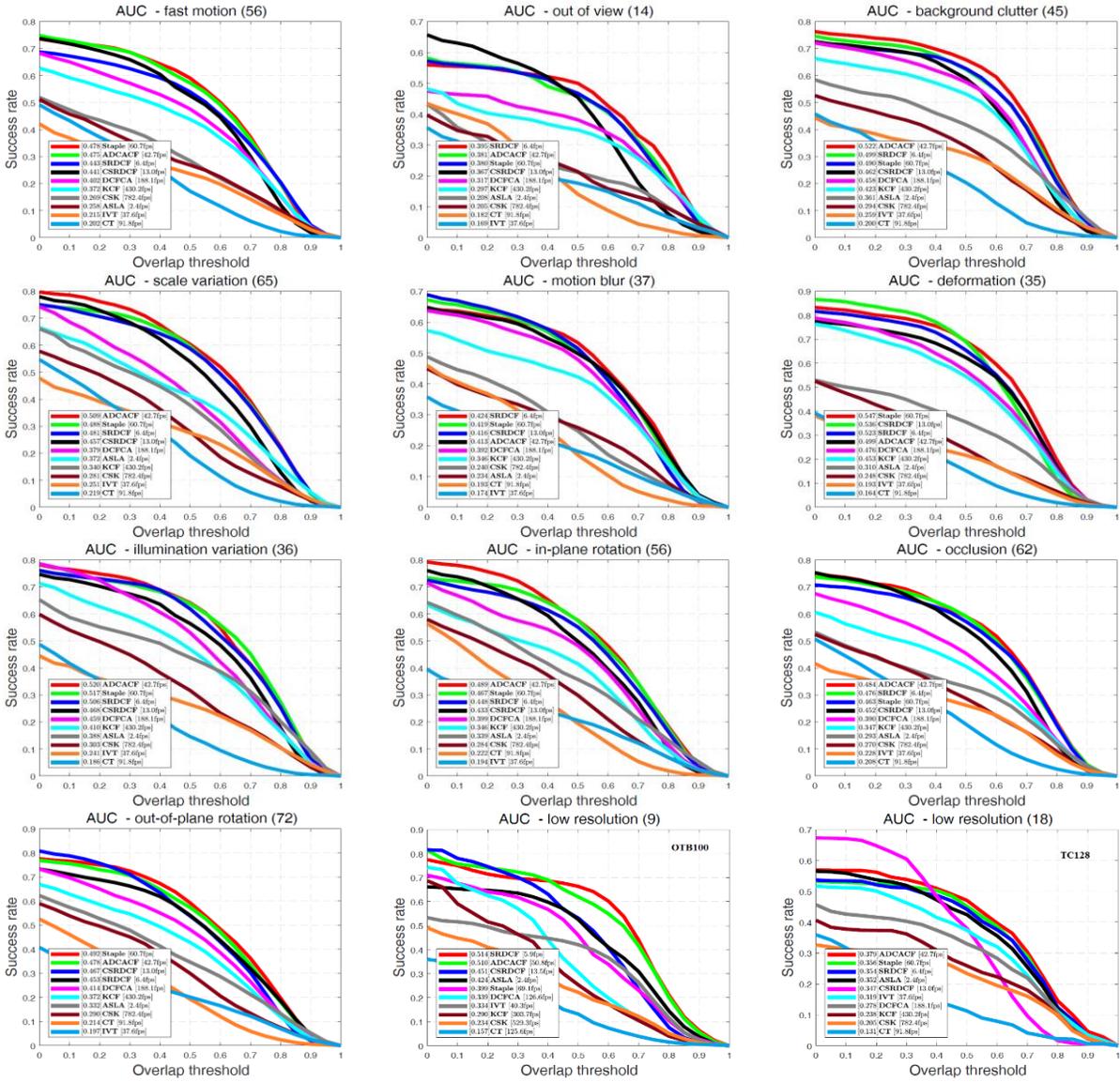